\documentclass[journal]{IEEEtran}
\usepackage{tikz}
\usepackage{cite}
\usetikzlibrary{calc}
\usetikzlibrary{arrows,shapes,chains}
\usepackage{xcolor}
\usepackage{amsfonts}
\usepackage{amsmath}
\usepackage{graphicx}
\usepackage{multirow}
\usepackage{algorithmic}
\usepackage{algorithm}
\usepackage{subfigure}
\usepackage{enumitem}
\usepackage{epstopdf}
\usepackage{url}
\usepackage{bm}

\usepackage{hyperref}
\usepackage{cleveref}
\crefname{section}{Section}{Sections}
\usepackage{amsthm,amsmath,amssymb}
\usepackage{mathrsfs}
\crefname{equation}{}{}
\crefname{table}{TABLE}{TABLES}
\crefname{figure}{Fig.}{Figs.}

\newcommand{\cb}[1]{\ifmmode {\boldsymbol{#1}}\else ${\boldsymbol{#1}}$\fi}

\newcommand{\bx}{\cb{x}}
\newcommand{\bz}{\cb{z}}

\newcommand{\be}{\cb{e}}

\newcommand{\by}{\cb{y}}
\newcommand{\bw}{\cb{w}}
\newcommand{\bW}{\cb{W}}
\newcommand{\bh}{\cb{h}}
\newcommand{\bE}{\cb{E}}
\newcommand{\ba}{\cb{a}}
\newcommand{\bP}{\cb{P}}

\usepackage{overpic}
\onecolumn
\title{Nonlinear Hyperspectral Unmixing based on Multilinear Mixing Model using Convolutional Autoencoders}
\author{Tingting~Fang, Fei~Zhu, \IEEEmembership{Member,~IEEE}, and Jie~Chen, \IEEEmembership{Senior~Member,~IEEE}
    \thanks{The work was supported by the National Natural Science Foundation of China under Grant 61701337. ~\emph{(Corresponding author: Fei Zhu)}}
	\thanks{T.~Fang and F.~Zhu are with the Center for Applied Mathematics, Tianjin University, China. (e-mail: fangtingts@163.com; fei.zhu@tju.edu.cn) }
	\thanks{J.~Chen is with the School of Marine Science and Technology, Northwestern Polytechnical University, Xi’an, China. (e-mail: dr.jie.chen@ieee.org).}
}

\begin{document}	
\maketitle
\begin{abstract}
Unsupervised spectral unmixing consists of representing each observed pixel as a combination of several pure materials called endmembers with their corresponding abundance fractions.
Beyond the linear assumption, various nonlinear unmixing models have been proposed, with the associated optimization problems solved either by traditional optimization algorithms or deep learning techniques.  
Current deep learning-based nonlinear unmixing focuses on the models in additive, bilinear-based formulations. 
By interpreting the reflection process using the discrete Markov chain, the multilinear mixing model (MLM) successfully accounts for the up to infinite-order interactions between endmembers.
However, to simulate the physics process of MLM by neural networks explicitly is a challenging problem that has not been approached by far. 
In this article, we propose a novel autoencoder-based network for 
unsupervised unmixing based on MLM.
Benefitting from an elaborate network design, the relationships among all the model parameters {\em i.e.}, endmembers, abundances, and transition probability parameters are explicitly modeled.
There are two modes: MLM-1DAE considers only pixel-wise spectral information, and MLM-3DAE exploits the spectral-spatial correlations within input patches.  
Experiments on both the synthetic and real datasets demonstrate the effectiveness of the proposed method as it achieves competitive performance to the classic solutions of MLM.

\end{abstract}

\section{Introduction}

%
\IEEEPARstart{S}{pectral} unmixing (SU) has been a hot topic in many fields related to remote sensing, {\em e.g.}, mineral exploration and agriculture over the past decades~\cite{bioucas2012hyperspectral}.
A hyperspectral image is a three-dimensional data cube, with hundreds of continuous spectral bands captured across a certain wavelength range over the same geographic target. 
Due to the relatively low spatial resolution, each spectra pixel in the image is supposed to be a mixture of several pure material signatures, namely {\em endmembers}. SU refers to identifying the endmembers and to estimating the corresponding fractions for each pixel, namely {\em abundances}~\cite{NonLU}. 

According to different assumptions on the photons' physical propagation process, the unmixing models are categorized into linear-based and nonlinear-based ones.~\cite{Heylen2014ReviewNonlinear}. Thanks to the simplicity of both the mixing process and unmixing strategy design, the linear mixing model (LMM) is the most prevalent. It assumes that a photon interacts with only one material before reaching the sensors, such that each observed spectrum is a linear combination of the endmembers ~\cite{bioucas2012hyperspectral}. 

However, severe nonlinear effects occur in many scenarios, thus requiring more sophisticated unmixing models~\cite{Heylen2014}. 
To this end, numerous nonlinear methods have been developed from the view of the physical mixing process, including the Hapke's model for intimate unmixing process~\cite{hapke1981bidirectional}, the bilinear-based methods~\cite{fan2009comparative, GBM2011, ppnm2012}, and the multilinear-based methods~\cite{Heylen2016, QiWei2017, yangbin2018, zhufei2019}.
The bilinear mixing models assume that each photon interacts with two endmember materials in series, thus mathematically formulated as the LMM superimposed by a second-order term.
By including different constraints on model parameters, representative bilinear-based unmixing models are Fan model (FM)~\cite{fan2009comparative}, generalized bilinear model (GBM)~\cite{GBM2011} and polynomial postnonlinear mixing model (PPNMM)~\cite{ppnm2012}. In~\cite{yang2022supervised},  a supervised SU method is developed for the bilinear mechanism with automatic shadow compensation. 

Recently, the multilinear mixing model (MLM) has been proposed to take account of all degrees of the endmember interactions, where a pixel-wise transition probability parameter is introduced to characterize the possibility of the light undergoing further interactions~\cite{Heylen2016}. 
The original MLM is established in a supervised manner, where the endmembers are supposed to be known in prior. It is regarded as the extension of the PPNMM from modeling the second-order nonlinear effects to the infinite order. 
After that, improvements have been made to MLM from different aspects. 
In~\cite{QiWei2017}, the so-called MLMp is formulated in an unsupervised manner to jointly estimate the endmembers, the abundance vectors, and the transition probability parameters.
This method employs a simplified optimization function to mitigate the parameter under-estimation issue and to reduce the algorithm complexity. 
The band-wise nonlinear unmixing algorithm (BNLSU)~\cite{yangbin2018} characterizes the transition probability parameter by a pixel-wise vector defined over the wavelength range instead of using a single scalar. 
In~\cite{zhufei2019}, a graph regularized multilinear mixing model is proposed, where the underlying manifold structure of the data is incorporated by using the Laplacian graph regularizers.
To enhance the robustness against noise suppress, 
A spectral-spatial reweighted robust unmixing method is proposed in~\cite{9955591}. This work enhances the robustness of the BNLSU by applying a $l_{2,1}$-norm-based loss function. 
Note that all the above variants of MLM are solved by traditional optimization methods, such as the alternating direction method of multipliers (ADMM). Different from previous works, we focus on addressing MLM by deep learning method in this paper.

Taking advantage of the progress in deep learning, the autoencoders (AEs) and their variants have been successfully applied to SU, especially for the linear-based unmixing problem~\cite{chen2022integration}.
The AEs are with an encoder-decoder architecture, in which neural networks are employed to find the low-dimensional representation of the input data by minimizing the reconstruction error~\cite{autoencoder}. 
When applying in SU, the encoder converts the input spectra to the corresponding abundance vectors,  {\em i.e.}, the outputs of the hidden layer, and the decoder reconstructs the input from the compressed representations, with the weights in the last linear layer interpreted as the endmember matrix. 
The majority of the deep learning-based unmixing strategies are based on the linear mixture assumption.  
The conventional AEs were employed for LMM in~\cite{Palsson2021Conv, Khajehrayeni2020, Menna2020, Palsson2018}. 
To mitigate the effects of noise in hyperspectral data, robust unmixing methods were developed, including the stacked nonnegative sparse autoencoders~\cite{su2018stacked} and the part-based denoising autoencoder with $l_{2,1}$ norm and denoising constraints~\cite{8126931}. 
The untied Denoising Autoencoder with Sparsity (uDAS) algorithm~\cite{qu2018udas} improves the part-based autoencoders by untying the encoder and decoder, and imposing the nonnegativity to only the decoder. 
Regularization terms were integrated to the loss function of AEs from different aspects, {\em e.g.}, the minimum simplex volume penalty to incorporate the geometry among data points~\cite{rasti2022misicnet,su2019daen,su2018stacked}, and the sparsity augmentation term on the abundance vectors~\cite{Dou2020,ozkan2018endnet}. 
The global-local smooth autoencoder (GLA) smooths the local homogeneity and global self-similarity of the image to better explore the spectral-spatial information~\cite{xu2022deep}.

Addressing the nonlinear SU in the context of deep learning is a challenging problem. It is because that nonlinear SU models usually require explicit descriptions of photons' interactions with the endmembers, while the black box characteristics of neural networks may lead to unclear mixing process with low interpretability~\cite{chen2022integration}. 
A category of nonlinear SU methods by deep learning is data-driven, and the nonlinearity is generally formulated in the form of implicit functions. 
In~\cite{wangmou2019}, a deep autoencoder network is designed to imitate an implicit nonlinear transform to the linear mixture, and the nonlinear layer is added to the decoder accordingly.
Other existing nonlinear SU models are realized by augmenting LMM with an additive nonlinear mixture part, where the nonlinearity is modeled implicitly~\cite{3DCNN2021, zhao2021lstm}. Different networks are applied, {\em i.e.}, a 3D-CNN structure to capture the spectral-spatial information~\cite{3DCNN2021}, and an LSTM to enhance the spectral correlations~\cite{zhao2021lstm}.
Attempts have also been made to realize the physics-driven unmixing models explicitly.
Considering the bilinear mixing models, two deep autoencoders within a multi-task learning framework are combined, but with the extra computational burden~\cite{su2020deep}.
A recent work~\cite{BMPPNM2021} designs a flexible unmixing scheme accommodating versatile nonlinear models from bilinear-based to higher-degree interactions.
The proposed network is equipped with a cross-product layer to model the nonlinear effects that occurred between endmembers, as well as a kernelization layer to separate mixing pixels in high-dimensional feature space.  
However, this method takes the learned network parameters rather than the learned features, as the pixel-wise, nonlinear coefficients of an input image, thus lack generalization ability. 

As far as we know, there exists no unmixing scheme based on deep learning fully conforming to the multilinear mixture mechanism, that takes account of the nonlinearity effects between endmembers up to infinite order. It is also noteworthy that compared to the additive, bilinear-based models, explicitly reproducing the physical unmixing process of MLM by neural networks is more challenging. 
In this paper, we propose an unsupervised autoencoder network for the MLM mechanism, where the nonlinearity is modeled explicitly through the elaborate network design. The proposed network is physics-driven with good interpretability and is capable of jointly estimate all the model parameters, {\em i.e.}, the endmembers, abundance, and nonlinear transition probabilities.
The main contributions of this paper are summarized as follows.
\begin{itemize}
\item[1)]
{{Physics-driven unmixing with interpretability}}: 
We propose an autoencoder-based network specifically for the MLM unmixing process. 
To the best of our knowledge, it is the first network fitting to the MLM. 
The proposed network has good interpretability in physics, as it imitates the unmixing process of MLM explicitly. 
The mathematical relationships of endmembers, abundances, and transition probability in the MLM model are explicitly modeled by smart network design with operations on learned features and specific network layers.
\item[2)]
{{End-to-end network for unsupervised unmixing with MLM}}: 
The proposed network addresses the MLM-based unmixing task in an unsupervised manner, where all the model parameters, namely the endmember matrix, abundance vector, and transition probability are estimated concurrently. 
It is an end-to-end framework that directly outputs the learned abundance vector and transition probability for each input data batch.

\item[3)]
{{Spectral-spatial unmixing network for MLM}}: 
According to whether exploiting the spatial information, the proposed unmixing network has two modes: MLM-1DAE which accounts for only spectral correlations, and MLM-3DAE which applies 3D convolution on image patches to incorporate the spectral-spatial information among the neighboring pixels. 

\end{itemize}

The remainder of the paper is organized as follows.
\cref{sec2} revisits the relevant unmixing models.
In~\cref{sec3}, the proposed autoencoder-based unmixing network for MLM is presented. 
Experimental results and analysis are given in~\cref{sec4}. Finally,~\cref{sec5} concludes the paper.

\section{Preliminary} \label{sec2}
Following are the notations that we use throughout the paper. Consider a hyperspectral image containing $N$ pixels, with $\boldsymbol E=[\boldsymbol e_1,\boldsymbol e_2,...,\boldsymbol e_R]\in\mathbb R^{B\times R}$ being the underlying endmember matrix composed by $R$ signatures over $B$ spectral bands. 
Let $\bx \in \mathbb{R}^{{B}\times{1}}$ be an observed pixel, $\boldsymbol a \in \mathbb{R}^{{R}\times{1}}$ be the associated abundance vector of the $R$ endmembers, and $\varepsilon \in \mathbb R^{B\times 1}$ is the additive Gaussian noise. 
The LMM considers to represent each pixel as a linear combination of the endmembers, and is formulated by 
\begin{equation}
\begin{split}
&\boldsymbol x=\boldsymbol E\boldsymbol a+\varepsilon,\\
&s.t.~a_i \ge 0,~\forall i=1,2,...,R\\
&~~~~~\boldsymbol 1_R^{\top}\boldsymbol a =1, \\
\end{split}
\end{equation}
where the non-negative constraint $($ANC$)$ and abundance sum-to-one constraint $($ASC$)$ are usually imposed considering the physical meaning. 

\subsection{Multilinear Mixing Model (MLM)~\cite{Heylen2016} }
In~\cite{Heylen2016}, the MLM extends the bilinear-based unmixing models to include all orders of interactions among endmembers. 
Based on the discrete Markov chain interpretation of the reflection process, the MLM assumes the following:
\begin{itemize}
\item  A photon will interact with at least one endmember material.
\item  After each interaction with an endmember, the photon has a probability $P$ for further interactions, and a probability $(1-P)$ of escaping the scenario and being received by the sensors.
\item  The probability of interacting with the $i$-th endmember is proportional to its corresponding abundance fraction.
\item  When scattered by some endmember, the light intensity will change according to that material's albedo $\bw \in \left[0, 1\right]^{B}$. By neglecting the nuance between the albedo and the endmember spectrum, the former is approximated by the latter with $\bw_i \approx \be_i$, $\forall i=1,2,...,R.$
\end{itemize}
Based on the above assumptions, the MLM is expressed by
\begin{equation}\label{mlm}
\begin{split}
{\bx} =&(1-P) \sum \limits_{i=1}^{R} \be_{i} a_{i}+ (1-P) P \sum \limits_{i=1}^{R} \sum \limits_{k=1}^{R}(\be_{i}\odot \be_{k}) a_{i} a_{k}\\
& +(1-P)P^{2}\sum \limits_{i=1}^{R} \sum \limits_{k=1}^{R} \sum \limits_{l=1}^{R}(\be_{i} \odot \be_{k}\odot \be_{l})a_{i}a_{k}a_{l}
 +\ldots +\varepsilon\\
= &(1-P)\by+ P\by \odot \bx+\varepsilon,
\end{split}
\end{equation}
where $\by=  \bE \ba$ represents the linear part with LMM model, and ${\odot}$ is the Hadamard element-wise product. 
Here, $P$ is the transition probability for further interactions, with its value restricted to $P<1$. When $P=0$, the model~\eqref{mlm} will degrade to the LMM. 
In theory, the above model is also well-defined for $P<0$, although the physical reasoning no longer holds with negative probability~\cite{Heylen2016}.
From~$\eqref{mlm}$, it is straightforward that 
\begin{equation}\label{xhat}
\begin{split}
\boldsymbol x = \frac{(1-P) \boldsymbol y}{1-P \boldsymbol y}+\varepsilon.
\end{split}
\end{equation}
As a result, the following optimization problem is considered, 
\begin{equation}\label{eq: optimization1}
\begin{split}
& \mathop{\arg\min}_{\boldsymbol a,P} \left\| \boldsymbol x - \frac{(1-P) \boldsymbol y}{1-P \boldsymbol y}  \right\|_2^{2} \\
& s.t.\quad\boldsymbol a\geq 0 \quad \text{and} \quad \cb{1}_R^{\top}\boldsymbol a=1\\
& \quad \quad~ P < 1, \\
\end{split}
\end{equation}
where the reconstruction error between the observed pixel and the reconstructed one is minimized. 
The unmixing problem in~\eqref{eq: optimization1} is formulated in a supervised manner, namely, the endmembers are supposed to be extracted in prior using some endmemeber extraction strategy, such as the vertex component analysis (VCA) algorithm~\cite{nascimento2005vertex}.

\subsection{Unsupervised MLM (MLMp)~\cite{QiWei2017}}
Later in~\cite{QiWei2017}, the authors proposed to address the MLM in~\eqref{mlm} as an unsupervised unmixing problem, where the endmember matrix, the abundance vector, and the transition probability parameter were estimated collaboratively.  
To this end, the reconstruction errors over all the pixels are added up, yielding 
\begin{equation}\label{eq: optimization2}
\begin{split}
& \mathop{\arg\min}_{\boldsymbol E, \{\boldsymbol a_j , P_j\}_{j=1}^N} \sum_{j=1}^N \left\|(1-P_j)\boldsymbol y_j+ P_j \boldsymbol y_j \odot \boldsymbol x_j -\boldsymbol x_j \right\|_2^{2} \\
& s.t.\quad \boldsymbol a_j \geq 0 \quad \text{and} \quad \boldsymbol {1}_R^{\top} \boldsymbol a_j=1\\
& \quad \quad \ P_j \leq 1, ~~\forall j=1,2,...,N \\
& \quad \quad \ 0 \leq \boldsymbol E \leq 1.
\end{split}
\end{equation}
To achieve the unsupervised unmixing with modest computational complexity, the above optimization simplifies the original problem~\eqref{eq: optimization1} by eliminating the denominator and applying a block coordinate descent method for problem-solving.
%

\section{Proposed method} \label{sec3}
\begin{figure*}  
\centering  
\graphicspath{{Figures/}}
\includegraphics[trim =0mm 0mm 0mm 0mm, clip,width=0.9\textwidth]{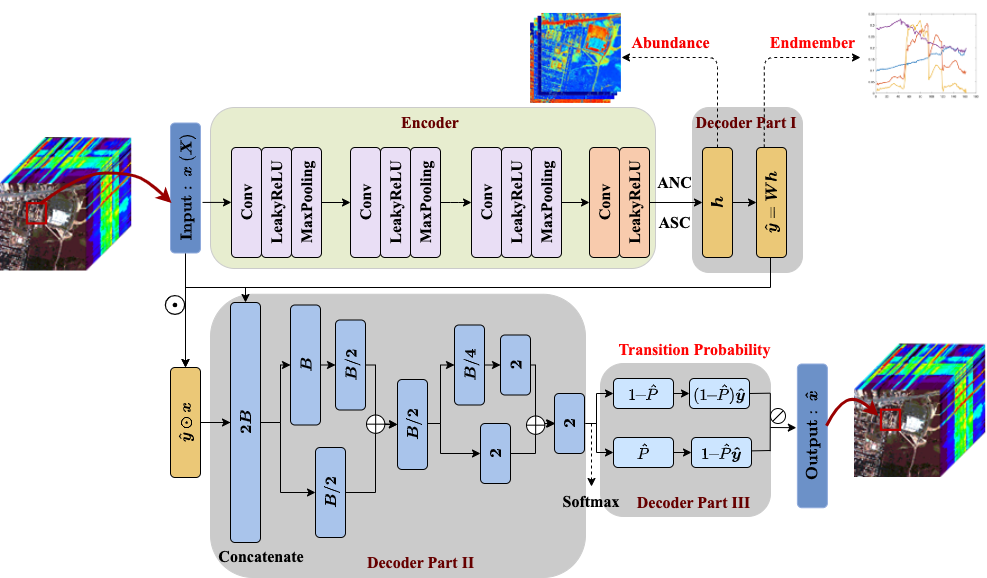}  
\caption{Schematic of the proposed autoencoder network for MLM-based unsupervised unmixing. 
}  
\label{main_net}  
\end{figure*}

\begin{figure*}  
\centering  
\graphicspath{{Figures/}}
\includegraphics[trim = 0mm 0mm 0mm 0mm, clip,width=0.92\textwidth]{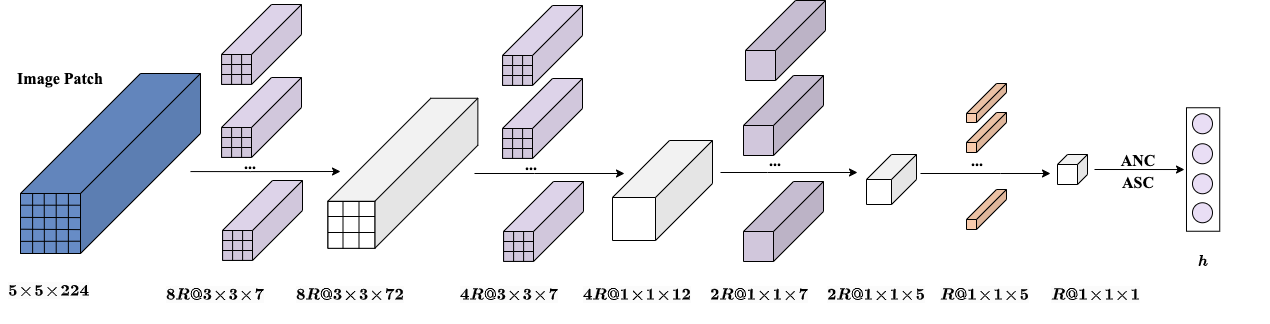}  
\caption{Illustration of feature changes during the encoding process in MLM-3DAE unmixing network, with $s=5, B=224$.
}  
\label{3DEncoder}  
\end{figure*}
In this section, we propose to address the MLM-based unmixing model in~\eqref{xhat} in an unsupervised manner, by exploiting an autoencoder-based network.
The proposed network directly simulates the physical process of MLM and is capable to estimate the endmembers, abundance vectors, and pixel-wise, transition probability parameters jointly. 
\subsection{Network overview}
The whole network is illustrated in~\figurename~\ref {main_net}.
It is based on an encoder-decoder architecture.
The encoder network mainly compresses the input pixel into a low-dimensional representation to obtain the abundance vector. 
The decoder network is more complex, which successively imitates the linear part with the LMM and built upon it, the MLM mechanism.    
Specifically, the linear part is performed by one linear layer in the decoder, with its network weights interpreted as the endmember matrix.
The mixture with the MLM model is simulated by reconstructing the pixel from model parameters according to~\eqref{xhat}, with operations such as feature reuse and concatenation.  
To determine the pixel-wise, transition probability $P$, which is subject to be within the interval [0,1] in this paper, features are concatenated, transformed, and finally pass a softmax layer to generate a probability score vector with $[P, 1-P]$.
The entire network is trained by minimizing the averaged distance between the input pixels and their reconstructions. 

\subsection{Encoder}
The encoder network $f(\cdot)$, which is shown in the upper part of~\figurename~\ref {main_net}, compresses the input data into the $R$-dimensional representation. 
Given the pixel $\bx$, the corresponding output of the encoder is taken as the estimated abundance vector $\hat{\boldsymbol a}$, namely 
\begin{equation}
{\text{Abundance~Estimation}}:~
{\boldsymbol a} ~ \leftarrow {\boldsymbol h} = f(\boldsymbol x)
\end{equation}
To enforce both the ANC and ASC constraints, the output feature of the encoder is normalized as follows
\begin{equation}\label{ASC}
h_i = \text{Softmax}(h_i)=\frac{e^{ h_i} }{\sum_{j=1}^R{e^{h_j}}}, 
\end{equation}
where $h_i$ denotes the $i$-th entry of the estimated abundance vector $\boldsymbol h$.
According to whether considering the spatial information, we employ two encoders with slight differences.

\subsubsection{MLM-1DAE}
The first encoder takes the pixel spectrum $\boldsymbol{x}$ as input, and 1D convolutions operations are performed across the spectral dimension in the subsequence blocks. 
Unmixing with this encoder is consistent with the original MLM~\cite{Heylen2016}, where only the spectral information is exploited during the unmixing process. 

To be precise, the first layer inputs a pixel of size $1\times 1\times B$.
Three identical blocks, each including a Convolution layer, a LeakyRelu layer and a MaxPooling layer are followed, and 
the fourth block has a convolution layer and a LeakyRelu layer. 
Regarding the convolution layers, the number of feature maps are $8R$, $4R$, $2R$, and $R$, and the kernel sizes are  $7$, $7$, $7$, and $5$, respectively, where no padding is added and all strides are set to $1$. 
For the MaxPooling layer, the kernel size is $3$ and the stride is 3, with no padding.
 We refer to the unmixing network using the encoder with 1D convolutions as MLM-1DAE. 

\subsubsection{MLM-3DAE}
To explore the rich spectral-spatial information of the hyperspectral imagery, we design the second encoder and refer to the unmixing network with it as MLM-3DAE. 
This encoder has a similar network structure as in MLM-1DAE and mainly utilizes the 3D convolutional layers to capture the spectral-spatial dependencies. 
The input becomes the image patch $\boldsymbol{X}$ of size $s \times s \times B$, which is centering at pixel $\boldsymbol{x}$. 
The aim is to extract the feature of size $1 \times 1 \times R$ from $\boldsymbol{X}$, to form the abundance vector for the neighborhood center 
$\boldsymbol{x}$. 
To downsample the spatial dimensions from $s\times s$ to $1\times 1$, we set the convolution kernel size in corresponding dimensions by $\max(3, odd(\lceil s/3 \rceil)) \times \max(3, odd(\lceil s/3 \rceil))$ without padding.
By the Convolution and MaxPooling layers over the spectral dimension, this dimension is decreased from $B$ to $1$, then R feature maps are concatenated to form the abundance vector.

For illustration purposes, the case with $s=5$ and $B=224$ is depicted in \figurename~\ref {3DEncoder}.
In the first block, the convolution layer has $8R$ feature maps with the 3D convolutional kernels of size $3\times 3\times 7$, no padding and $1\times 1\times 1$ stride; the MaxPooling layer has the kernel size $1\times 1\times 3$, with no padding and $1\times 1\times 1$ stride. 
The second block has $4R$ feature maps, with other settings unchanged. 
After this block, the spatial dimensions become $1\times 1$.
The third block has $2R$ feature maps with 3D convolution kernels of size $1\times 1\times 7$, no padding and stride size $1\times 1\times 1$; the kernel size in MaxPooling layer is $1\times 1\times 3$.  
The fourth block has $R$ feature maps which corresponds to the number of endmember, and the 3D convolution kernel is of size $1\times 1\times 5$ without padding.

\subsection{Decoder} 
In this paper, the decoder network mimics the linear part with the LMM and based upon it, the nonlinear mixture mechanism with the MLM model. The encoded feature dimension is mapped from $R$ to $B$ to reconstruct the input. 
Our decoder is composed by three components, as shown in \figurename~\ref {main_net}.
\subsubsection{Decoder Part I}
The first layer of the decoder is the fully connected (FC) layer without bias and activation function, following the last layer of the encoder. 
It processes the linear transformation on the encoder outputs, {\em i.e.}, abundance vectors, and is taken as the linear mixture of the LMM model. 
The network weights $\boldsymbol W$ of the FC layer is interpreted as the endmember matrix by
\begin{center}
Endmember~Extraction:~$\boldsymbol E~ \leftarrow \boldsymbol W$.
\end{center}
Here, the layer weights are restricted to the interval [0,1] by applying $\phi(x)=\min(1,\max(0,x))$, considering the physical meaning of endmembers. 
In practice, we initialize the weights of this FC layer using the endmembers extracted by the VCA technique~\cite{nascimento2005vertex}.

From the MLM model in~\eqref{mlm}, the observed pixel can be taken as a weighted sum of a linear component ${\boldsymbol y}$ and a nonlinear part ${\boldsymbol y}\odot {\boldsymbol x}$, with the proportion between them adjusted by the transition probability parameter $P$. 
To this end, we concatenate the two features, which are the output of precedent layer (the linear part ${\hat{\boldsymbol y}}=\bW \bh$), and its Hadamard product with the input data given by $\hat{\boldsymbol y}\odot {\boldsymbol x}$. The resulting feature, which is of an enlarged dimension $2 B$, is then fed to the second part of the decoder.

\subsubsection{Decoder Part II}
The second part is a deep neural network with skip-connections (DNNSc),  which is elaborated to extract the high-level information related to the transition probability parameter from the concatenated feature. It is composed by a set of FC layers, hyperbolic tangent function (Tanh) activation layers and two skip-connections,  and gradually reduces the feature dimension from $2B$ to 2. 

We apply two skip connections to alleviate the gradient vanishing when the network becomes deep~\cite{He_2016_CVPR}. It also helps to improve the unmixing performance of the proposed network.
Following the notations in~\cite{He_2016_CVPR, he2016identity}, the skip connection layer is represented as 

\begin{equation}
c_{l+1}=\Psi({Q_1c_{l}})+\Psi({Q_2c_{l}}),
\end{equation}
where $c_l$ and $c_{l+1}$ represent the layer input and output, $Q_i$ is a linear mapping, and $\Psi$ is the activation function.

To reduce the feature dimension from $2B$ to $B/2$, two paths are separated and then joint up.  
The first path transforms the dimension from $2B$ to $B$ by a linear layer and an activation layer; and from $B$ to $B/2$ by a linear layer.
The second path applies the skip\- connection that directly maps the feature dimension from $B$ to $B/2$.
The two paths are added up, followed by an activation function.
The subsequent layers are similarly designed to further decrease the dimension to 2, but without applying the activation layers when the dimension is 2.

\subsubsection{Decoder Part III}
The last part of the decoder estimates the transition probability parameter by a softmax function, and reconstructs the spectrum according to~\eqref{xhat} for MLM model. 
The softmax function is applied to the output of the precedent layer, denoted by $\bz=[z_1, z_2]^\top$, and transforms it to the binomial probability distribution with ${\bP}=[1-\hat{P}, \hat{P}]^\top$.
As a result, the transition probability is estimated by 
\begin{equation}
{\text {Probability~Estimation}}:~ P \leftarrow \hat{P}=\frac {\exp z_2}{\sum_{i=1}^2 \exp z_i}
\end{equation}
In this manner, the estimated value of $P$ will be subject to the interval $[0,1]$, a reasonable range to describe the parameter as the probability of a photon undergoing further interactions in the reflectance process of the MLM model. 
By reusing the learned features and the network parameters,  the rest part of the decoder reconstructs the input data by the division operation between the feature ${(1-\hat{P}) \hat{\boldsymbol y}}$ and the feature ${(1-\hat{P} \hat{\boldsymbol y})}$, following the formula in~\eqref{xhat}.

\subsection{Loss Function}
We adopt the commonly-used spectral angle distance (SAD) between the observed pixels and reconstructed ones as the loss function of the proposed network.
The SAD measures the distance between two vectors, and is defined by 
\begin{equation} \label{sad}
J_{\text{SAD}}(\boldsymbol x,\hat {\boldsymbol x})=\arccos (\frac{ {\bx}^\top \hat{\bx}}{\| \bx \|_2 \|\hat{\bx}\|_2}),
\end{equation}
where $\hat{\boldsymbol x }= \frac{(1-\hat P) \hat{\boldsymbol y}}{1-\hat P \hat{\boldsymbol y}}$ is the reconstructed pixel obtained from the output of the network.
The loss function becomes
\begin{equation} 
\mathcal{L}=\frac{1}{K}\sum_{k=1}^K{J_{\text{SAD}}(\boldsymbol x_i,\hat {\boldsymbol x}_i)},
\end{equation}
where $K$ is the batch size. 

Notice we directly tackle the original loss function of MLM model in~\eqref{eq: optimization1}, despite replacing the Euclid distance with the cosine distance. 
It is different from the MLMp algorithm that adopts a simplified loss function as in~\eqref{eq: optimization2} for unsupervised unmixing.

\section{experiments}\label{sec4} 
In this section, we prove the effectiveness of the proposed multilinear-based unmixing networks MLM-1DAE and MLM-3DAE on a series of synthetic images, as well as two real hyperspectral images.
Experiments are implemented on NVIDIA GeForce RTX 2080 Ti, using the deep learning framework PyTorch.
For a fair comparison, five state-of-the-art unmixing methods are chosen by considering two aspects.  
They are two classic unmixing schemes developed based on the multilinear mixture assumption, in either a supervised or an unsupervised setting, and three well-known autoencoder-based networks proposed for linear and nonlinear unmixing.
\begin{itemize}
\item
VCA~\cite{nascimento2005vertex} + MLM~\cite{Heylen2016}: As the multilinear unmixing method MLM is supervised, we extract the endmembers by VCA in prior, and then apply the MLM to estimate the abundances and the transition probabilities.
\item
MLMp~\cite{QiWei2017}: The so-called MLMp is a unsupervised unmixing strategy based on  the multilinear mixture model, where all the model parameters including endmembers,  abundances and trancsition probabilities are estimated jointly by considering a simplified 
optimization problem.
\item 
uDAS~\cite{qu2018udas}: It is a denoising autoencoder developed for LMM, where the encoder and the decoder are untied and the denoising and sparsity constraints are considered. The uDAS is competitive for unmixing the highly noisy data.
\item
3D-NAE~\cite{3DCNN2021}:
This method addresses a general nonlinear unmixing model consisting of a linear mixture part and an additive nonlinear mixture part. An autoencoder-based network is designed to make full use of spectral and spatial information by using CNN. The nonlinearity is modeled by an implicit function.
\item
PPNM-AE~\cite{BMPPNM2021}:
This flexible autoencoder accommodates versatile nonlinear unmixing models, namely the postpolynomial nonlinear mixing (PPNM) model, by employing a cross-product layer. 
A kernelization layer is designed to separate the mixing pixels in high-dimensional feature space for better parameter estimation. 
\end{itemize}
The endmembers are identically initialized in comparing methods with the signatures extracted by VCA,
except for PPNM-AE which uses k-means as endmember initialization.

\subsection{Synthetic Data}
\subsubsection{Data Generation}
A synthetic data of size $256\times256$ pixels is generated according to the MLM mixture mechanism given in~\eqref{mlm}. 
We choose $R=4$ groundtruth endmembers from the United States Geological Survey (USGS) digital spectral library composed by spectra measured over 224 consecutive bands. 
The abundance maps are generated using the Hyperspectral Imagery Synthesis(HYDRA) toolbox
\footnote{http://www.ehu.es/ccwintco/index.php/HyperspectraI Imagery Synthesis tools for MATLAB}, as shown in~\figurename~\ref{abunSNR30}. 
{The pixel-wise transition probabilities are randomly generated from a half-normal distribution with $\sigma=0.3$, and the values of $P$ larger than 1 are set to zero, as did in~\cite{Heylen2016}~\cite{QiWei2017}.}
In the experiments, we consider different noise levels by adding the Gaussian white noise with a signal-to-noise ratio (SNR) at 25, 30, and 35, respectively. 

\subsubsection{Evaluation Metrics}
As all the groundtruth is available, different methods are evaluated comprehensively in terms of all the model parameters, including endmembers, abundances, and transition probabilities, by using following criterions.

%

The endmember estimation is evaluated by using the averaged spectral angle distance (SAD) of all the endmembers, which is defined by 
\begin{equation}
\mathrm{SAD_{end}}=\frac{1}{R}\sum_{i=1}^R { \arccos}(\frac{\boldsymbol m_i ^\top \hat{\boldsymbol m_i }}{\left\|\boldsymbol m_i \right\|_2 \left\| \hat{\boldsymbol m_i } \right\|_2}), 
\end{equation}
where $\boldsymbol m_i$ and $\hat{\boldsymbol m_i}$ represent the $i$-th groundtruth and estimated endmembers, respectively. 

The abundance estimation is evaluated using the root mean square error (RMSE), given by
\begin{equation}
\mathrm{RMSE_{abun}}=\sqrt{\frac{1}{NR}\sum_{i=1}^N{\left\|{\boldsymbol{a}_i}-\hat{\boldsymbol{a}}_i\right\|}_2^2}
\end{equation}
where $\boldsymbol a_i$ and $\hat{\boldsymbol{a}_i}$ are the groundturth and estimated abundance vectors for the $i$-th pixel.

The transition probabilities obtained by the multilinear-based unmixing methods are compared by the RMSE, with
\begin{equation}
\mathrm{RMSE_P}=\sqrt{\frac{1}{N}\sum_{i=1}^N{\left\|P_i-\hat{P}_i\right\|}^2}
\end{equation}
where $P_i$ and $\hat{P}_i$ are the groundtruth and the estimated values, respectively.

Finally, we consider the averaged pixel SAD between the observed and reconstructed pixels, with the following expression 
\begin{equation}
\mathrm{SAD}={\frac{1}{N} \sum_{i=1}^N \arccos ( \frac{ {\bx}_i ^\top \hat{\bx}_i } {\| \bx_i  \|_2 \|\hat{\bx}_i\|_2})},
\end{equation}
where $\boldsymbol x_{i}$ and $\hat {\boldsymbol x}_{i}$  are the actual and reconstructed pixels.

\subsubsection{Paramter settings}
In this series of experiments, the hyperparameters in the proposed MLM-1DAE and MLM-3DAE are set as follows.
The Adam optimizer is adopted, where the learning rate strategy varies for different parts of the network. 
Specifically, the initial learning rate in the linear decoder is set to $0.0005$ with an exponential decay of $0.9$ after each epoch, while the rest parts of the network apply a constant learning rate of $0.001$. 
The number of training epochs is set to $150$.

\begin{figure} [t] 
\centering  
\graphicspath{{Figures/}}
\includegraphics[width=0.4\textwidth]{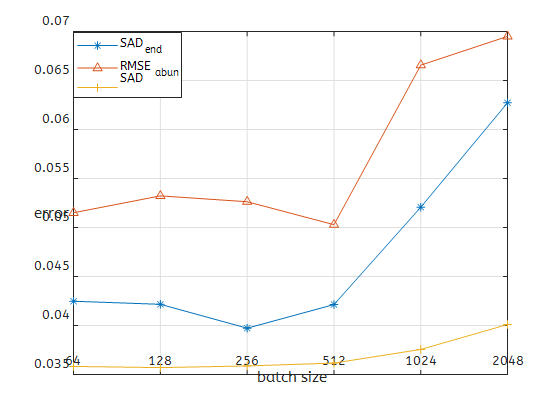}  
\caption{Average endmember SAD, abundance RMSE, and reconstruction error (RE) values as a function of the batch size using MLM-1DAE.}
\label{batch size}  
\end{figure}

We tune the batch-size from the candidate value set $[64, 128, 256, 512, 1024, 2048]$ using the MLM-1DAE method, on the synthetic image with SNR=30. \figurename~\ref{batch size} reports the resulting errors with respect to three monitoring metrics, {\em i.e.}, averaged endmember SAD, abundance RMSE, and averaged pixel SAD. Accordingly, we choose the batch size to be 512 in both MLM-1DAE and MLM-3DAE.

\begin{figure}  [t]
\centering  
\graphicspath{{Figures/}}
\includegraphics[width=0.4\textwidth]{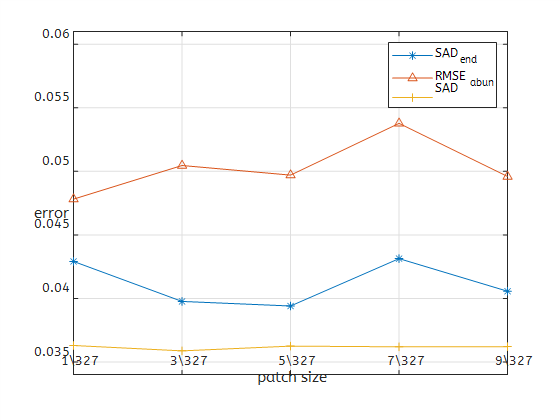}  
\caption{Changes of the averaged endmember SAD, abundance RMSE, and averaged pixel SAD along with the varying input patch size using MLM-3DAE.}  
\label{patch size}  
\end{figure}



Regarding MLM-3DAE, the patch size $s \!\times\! s$ for the input subcube is selected from the candidate set $[1\times 1, 3\times3, 5\times5, 7\times7, 9\times9]$, and the corresponding changes of the averaged endmember SAD, abundance RMSE, and averaged pixel SAD are given in~\figurename~\ref{patch size}. Here, we apply the MLM-1DAE with 1D convolution in $s \times s \!= \!1 \times 1$.
As observed, modest patch sizes with $3\times3$ and $5\times5$ can efficiently use the spatial information of neighboring pixels by 3D convolution operations, yielding better unmixing performance when compared to merely using the spectral information with $1\times1$. 
However, a greater spatial patch size with $7\times7$ will not only increase the computing overhead, but also deteriorate the unmixing performance. It is because of the overfitting issue caused by an increasing number of network parameters. 
Based on the above analysis, we choose the patch size in MLM-3DAE to be $5\times5$ in the following experiments. 

\subsubsection{Results Analysis}

\begin{table*}[htbp]\caption{Comparison of the averaged endmember SAD $\times 10^{-2}$ (mean $\pm$ standard deviation) on synthetic data at SNR$=$25,30,35, averaged over 5 runs.}
	\centering
     
	\begin{tabular}{c|c|c|c|c|c|c|c}
		\hline
		\hline
		                  &$$VCA+MLM$$    &$$MLMp$$  &$$uDAS$$  &$$3D-NAE$$ &$$PPNM-AE$$ &$$MLM-1DAE$$  &$$MLM-3DAE$$\\
		\hline
		
		$$SNR=25dB$$   &$10.41\pm0.00$  &$19.76\pm0.00$  &$11.87\pm0.00$    &$12.54\pm0.08$ &$27.06\pm0.00$	 &$\textcircled{\tiny1}~{9.32\pm0.08}$ &$\textcircled{\tiny2}~9.34\pm0.59$\\
		$$SNR=30dB$$   &$8.71\pm0.00$	&$4.97\pm0.00$  &$9.75\pm0.00$  &$11.02\pm0.28$ &$ 26.12\pm0.00$ &$  \textcircled{\tiny2}~4.42\pm0.22$  &$\textcircled{\tiny1}~ {3.98\pm0.29}$ \\
		$$SNR=35dB$$  &$~5.96\pm0.00$	&$\textcircled{\tiny1}~{2.93\pm0.00}$  &$5.88\pm0.00$  &$7.55\pm0.10$&$25.75\pm0.00$  &$3.64\pm0.45$&$\textcircled{\tiny2}3.54\pm0.53$          \\		
		\hline
		\hline
	\end{tabular}
\label{endmember_SAD}
\end{table*}

\begin{table*}\caption{Comparison of the abundance RMSE $\times 10^{-2}$ (mean $\pm$ standard deviation) on synthetic data at SNR$=$25,30,35, averaged over 5 runs.}
	\centering
	
	\begin{tabular}{c|c|c|c|c|c|c|c}
	        \hline
		\hline
		      &$$VCA+MLM$$ &$$MLMp$$ &$$uDAS$$  &$$3D-NAE$$ &$$PPNM-AE$$ &$$MLM-1DAE$$  &$$MLM-3DAE$$\\
		\hline		
		$$SNR=25dB$$     &$7.01\pm0.00$  &$ 6.99\pm0.00$  &$10.08\pm0.00$&$12.09\pm0.10$&$14.73\pm0.00$	&$\textcircled{\tiny1}~5.86\pm0.41$  &$\textcircled{\tiny2}~6.43\pm0.29$\\
		$$SNR=30dB$$  &$ 7.86\pm0.00$  &$5.73\pm0.00$  &$10.01\pm0.00$ &$12.63\pm0.11$&$15.69\pm0.00$&$\textcircled{\tiny2}~5.25\pm 0.34$&$ \textcircled{\tiny1}~4.99\pm 0.55$\\
		$$SNR=35dB$$ &$7.15\pm0.00$  &$6.98\pm0.00$ &$8.22\pm0.00$&$13.24\pm0.15$&$15.73\pm0.00$&$\textcircled{\tiny2}~{4.75\pm0.54}$&$ \textcircled{\tiny1}~{4.51\pm0.71}$\\
		
		\hline
		\hline
	\end{tabular}
\label{abundance_RMSE}
\end{table*}

\begin{table*}\caption{Comparison of the averaged pixel SAD $\times 10^{-2}$ (mean $\pm$ standard deviation) on synthetic data at SNR$=$25,30,35, averaged over 5 runs.}
	\centering
	
	\begin{tabular}{c|c|c|c|c|c|c|c}
		\hline
		\hline
		                       &$$VCA+MLM$$     &$$MLMp$$  &$$uDAS$$   &$$3D-NAE$$  &$$PPNM-AE$$ &$$MLM-1DAE$$  &$$MLM-3DAE$$\\
		\hline
		
		$$SNR=25dB$$   &$6.49\pm0.00$  &$\textcircled{\tiny2}~{6.41\pm0.00}$  &$7.83\pm0.00$   &$7.01\pm0.07$  &$8.6\pm0.00$&$\textcircled{\tiny1}~6.30\pm0.02$&$6.42\pm0.24$  \\
		$$SNR=30dB$$ &$4.59\pm0.00$  &$\textcircled{\tiny1}~{3.58\pm0.00}$  &$6.58\pm0.00$ &$4.60\pm0.03$&$6.92\pm0.00$&$3.62\pm0.01$&$\textcircled{\tiny2}~3.61\pm0.02$\\
		$$SNR=35dB$$ &$3.54\pm0.00$  &$2.89\pm0.00$  &$4.49\pm0.00$&$3.54\pm0.02$&$6.25\pm0.00$&$\textcircled{\tiny1}~2.12\pm0.02$&$\textcircled{\tiny2}~{2.12\pm0.03}
$\\
		\hline
		\hline
	\end{tabular}
\label{RE_sad}
\end{table*}

\begin{table*}\caption{Comparison of RMSE of P value $\times 10^{-2}$ (mean $\pm$ standard deviation) on synthetic data at SNR$=$25,30,35, averaged over 5 runs.}
	\centering	
	\begin{tabular}{c|c|c|c|c}
		\hline
		\hline
		                     &$$VCA+MLM$$     &$$MLMp$$    &$$MLM-1DAE$$  &$$MLM-3DAE$$\\
		\hline
		
		$$SNR=25dB$$ &$35.57\pm0.00$  &$56.9\pm0.00$  &$\textcircled{\tiny1}~9.87\pm2.76$   &$\textcircled{\tiny2}~{13.82\pm4.91}$  \\
		$$SNR=30dB$$  &$37.78\pm0.00$  &$\textcircled{\tiny1}~{12.30\pm0.00}$  &$28.25\pm2.45$ &$\textcircled{\tiny2}~27.08\pm0.82$\\
		$$SNR=35dB$$  &$\textcircled{\tiny2}~19.62\pm0.00$  &$\textcircled{\tiny1}~{10.57\pm0.00}$  &$26.48\pm5.41$&$25.12\pm6.07$\\
		
		\hline
		\hline
	\end{tabular}
\label{RMSEp}
\end{table*}

\begin{figure*}  
	\centering
       \graphicspath{{Figures/}}
	\subfigure[{End\#1}]{
		\label{a_end_1}
		\includegraphics[trim =6.5mm 0mm 13mm 0mm, clip,width=0.4\textwidth]{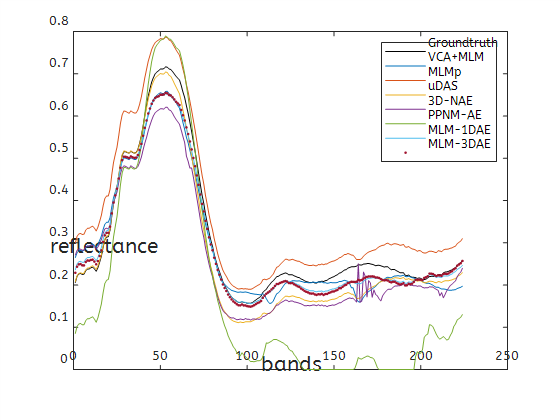} 
	}   
	\subfigure[End\#2]{
		\label{b_end_2}
		\includegraphics[trim =6.5mm 0mm 13mm 0mm, clip,width=0.4\textwidth]{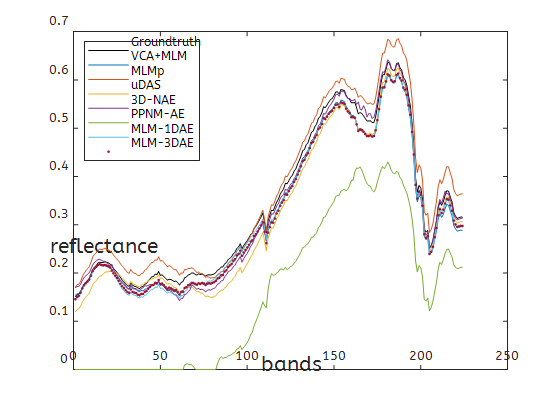}
        }
\\
	\subfigure[End\#3]{
		\label{c_end_3}
		\includegraphics[trim =6.5mm 0mm 13mm 0mm, clip,width=0.4\textwidth]{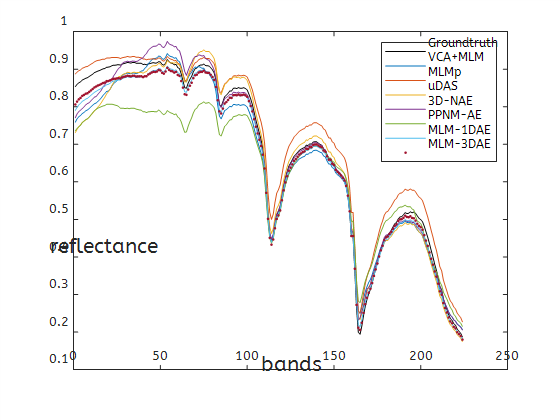} 
       }
	\subfigure[End\#4]{
		\label{d_end_4}
		\includegraphics[trim =6.5mm 0mm 13mm 0mm, clip,width=0.4\textwidth]{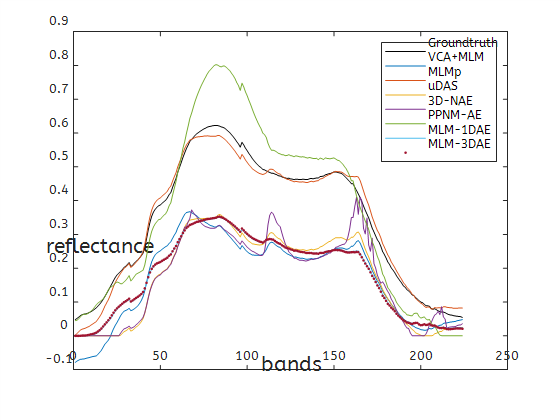}
       }
\caption{ Endmember extraction results of synthetic data at $SNR=30$.~$($a$)$~Endmember~1.~$($b$)$~Endmember~2.~$($c$)$~Endmember~3.~$($d$)$~Endmember~4.} 
\label{End_synthetic}
\end{figure*}

\begin{figure*}
\centering
\small  
\tabcolsep=0.1mm
\graphicspath{{Figures/}}
\begin{tabular}{ccccccccc}
Groundtruth & VCA+MLM&MLMp&uDAS&3D-NAE&PPNM-AE&MLM-1DAE&MLM-3DAE&~\\
\rotatebox{90}{ ~~~ End\#1~~~~~~~~ End\#2~~~~~~~~ End\#3~~~~~~~~~End\#4}
		\includegraphics[trim =3mm 0mm 0mm 0mm, clip,width=0.11\textwidth]{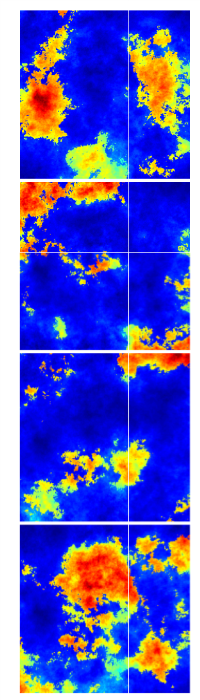} &
		\includegraphics[trim = 3mm 0mm 0mm 0mm, clip,width=0.11\textwidth]{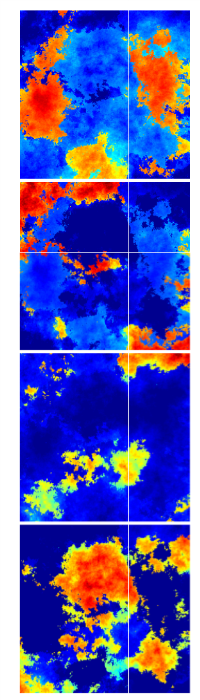}  &
             \includegraphics[trim = 3mm 0mm 0mm 0mm, clip,width=0.11\textwidth]{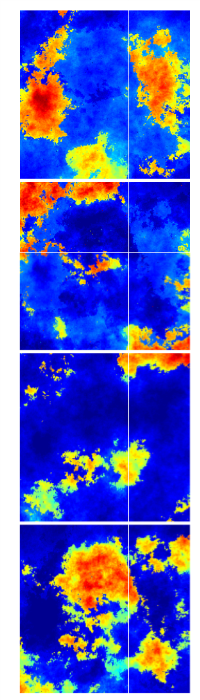}&
             \includegraphics[trim = 3mm 0mm 0mm 0mm, clip,width=0.11\textwidth]{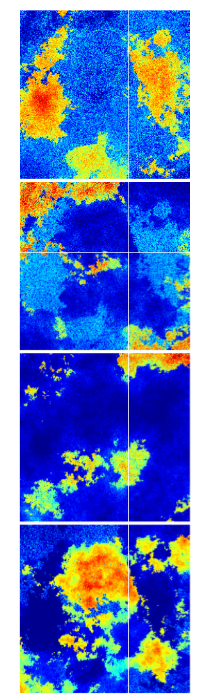} &
             \includegraphics[trim =3mm 0mm 0mm 0mm, clip,width=0.11\textwidth]{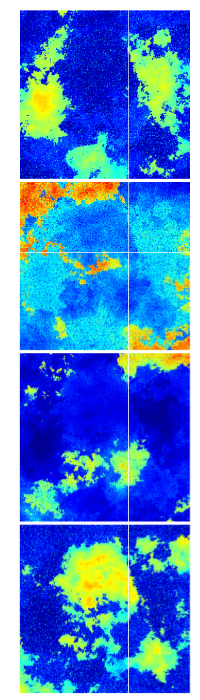} &
		\includegraphics[trim = 3mm 0mm 0mm 0mm, clip,width=0.11\textwidth]{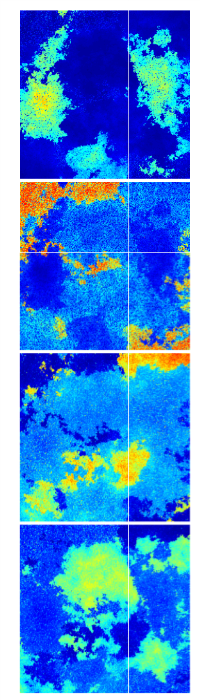}  &
             \includegraphics[trim = 3mm 0mm 0mm 0mm, clip,width=0.11\textwidth]{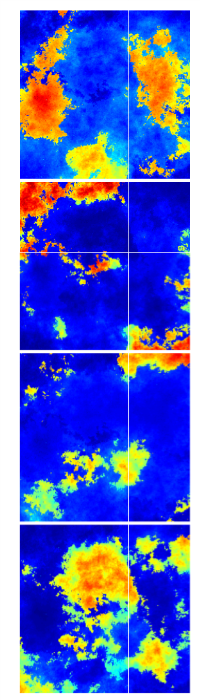}&
             \includegraphics[trim = 3mm 0mm 0mm 0mm, clip,width=0.11\textwidth]{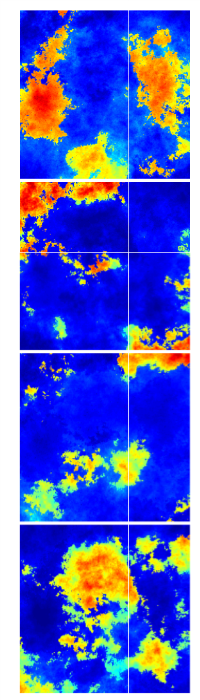} &
		\includegraphics[trim =35mm 19mm 0.8mm 8.5mm, clip,width=0.045\textwidth,height=7.55cm]{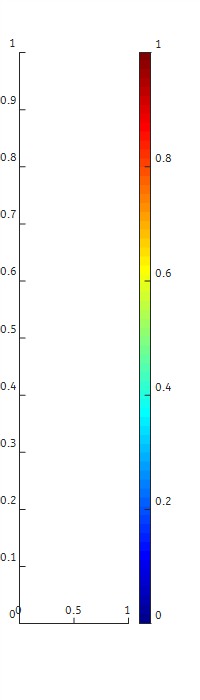} 
\end{tabular}
\caption{ Estimated abundance maps on synthetic data for all the four endmembers, at $SNR=30$. From left to right: Groundtruth, VCA+MLM, MLMp, uDAS, 3D-NAE, PPNM-AE, MLM-1DAE, and MLM-3DAE }  
\label{abunSNR30}  
\end{figure*}

\begin{figure*}  
	\centering  
\graphicspath{{Figures/}}
	\subfigure[Groundtruth]{
			\label{fig:subfig:A}
			\includegraphics[trim = 4mm 0mm 14mm 5mm, clip,width=0.28\textwidth]{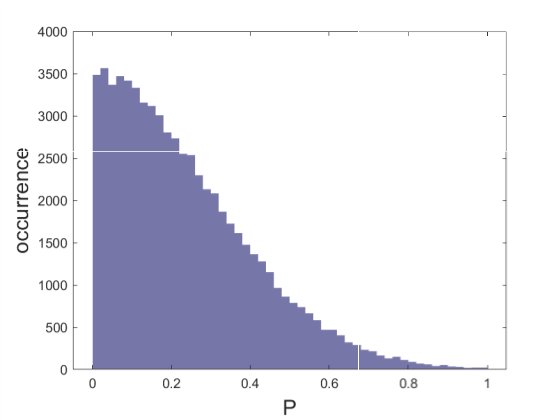}
	        }
	\subfigure[VCA+MLM]{
			\label{fig:subfig:A}
			\includegraphics[trim = 4mm 0mm 14mm 5mm, clip,width=0.28\textwidth]{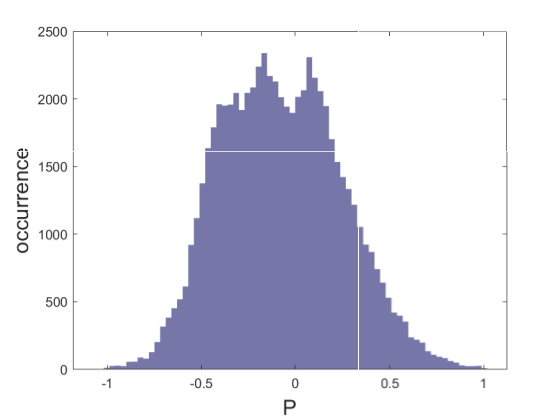}
	        }
\\
	\subfigure[MLMp]{.png
			\label{fig:subfig:b}
			\includegraphics[trim = 4mm 0mm 14mm 5mm, clip,width=0.28\textwidth]{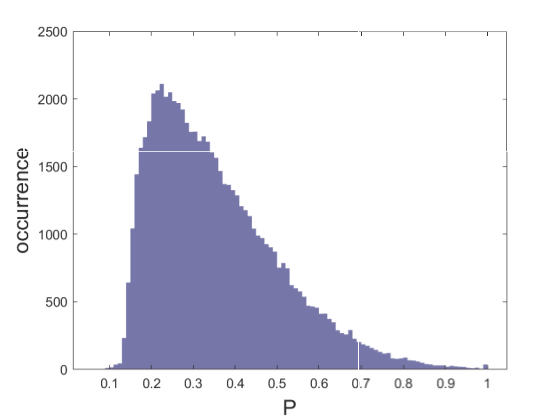}
	        }
    \subfigure[MLM-1DAE]{
			\label{fig:subfig:b}
			\includegraphics[trim = 4mm 0mm 14mm 5mm, clip,width=0.28\textwidth]{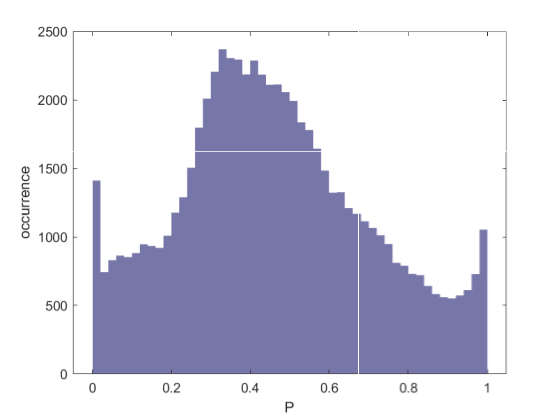}
			
	        }
  \subfigure[MLM-3DAE]{
			\label{fig:subfig:b}
			\includegraphics[trim = 4mm 0mm 14mm 5mm, clip,width=0.28\textwidth]{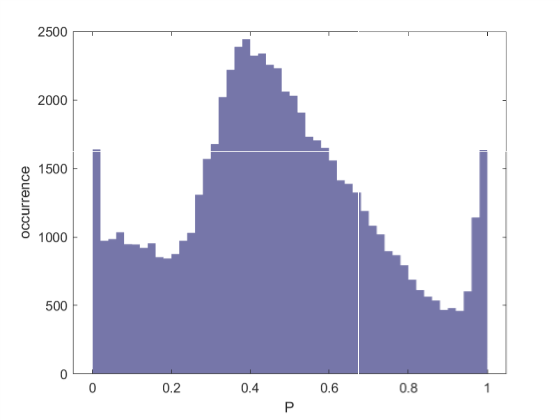}
			
	        }
	\vfill
\caption{Histograms of the transition probability $P$ obtained by using different methods on synthetic data at SNR=30.
}  
\label{p}  
\end{figure*}


All the experiments are repeated five times, and the unmixing results in terms of mean and deviation,   obtained by different unmixing methods are quantitively compared in \tablename~\ref{endmember_SAD}-\ref{RMSEp}. 
The best and the second-best results are marked by a circled number.
In general, the MLM-based unmixing strategies perform better than the comparing methods designed for other linear and nonlinear models, in all the comparing criteria. This phenomenon is reasonable as the unmixing performance is highly related to the consistency between the data mixing mechanism and unmixing model applied. 

Concerning the endmember extraction,  
\tablename~\ref{endmember_SAD} reports the averaged endmember SAD obtained by all the algorithms at varying noise levels, and \figurename~\ref{End_synthetic} illustrates the endmember extraction results of one run. 
We observe that the proposed MLM-1DAE and MLM-3DAE achieve the best two performances for the cases with SNR=25,30, but are inferior to the MLMp for the case with SNR=35. 
Of particular note is the classic MLMp is overwhelmingly advantageous in endmember estimation when the data is relatively clean with SNR=35. 
The three counterparts based on autoencoder networks fail to estimate the endmembers accurately mainly due to the mismatch between the mixing and unmixing models. 


The estimated abundance RMSEs achieved by different methods are reported in  \tablename~\ref{abundance_RMSE}, 
where the proposed MLM-1DAE and MLM-3DAE outperform other methods in this metric on all the three datasets. 
Specifically, for relatively clean data with SNR=30 and 35, MLM-3DAE yields better abundance estimation than MLM-1DAE, as the former takes advantage of the spectral-spatial dependencies contained in the imagery. 
\figurename~\ref{abunSNR30} visualizes the abundance maps obtained by different methods. 
Compared to the non-MLM unmixing counterparts, the MLM-based unmixing methods, including VCA+MLM, MLMp, and the proposed MLM-1DAE and MLM-3DAE, generally result in abundance maps visually more consistent with the groundtruth.

Table \ref{RE_sad} reports the averaged pixel SAD. 
At SNR=30, it is MLMp that leads to the best performance, second by the proposed MLM-3DAE. 
At SNR=35, the proposed MLM-1DAE results in the smallest SAD among all the comparing methods, followed by MLMp. 
This proves the effectiveness of the proposed methods in umixing the data generated by MLM, as comparable results to MLMp are obtained. 

We compare the transition probability RMSEs obtained by different MLM-based methods in~\tablename~\ref{RMSEp}. At SNR=30 and 35, MLMp leads to the best estimation of the transition probability, whereas at SNR=25, the proposed MLM-1DAE and MLM-3DAE are more competitive. 
The histogram of P is also shown in \figurename~\ref{p}.

\begin{figure*}
\graphicspath{{Figures/}}
\centering
\small  
\tabcolsep=0.1mm
\begin{tabular}{cccccccc}
VCA+MLM&MLMp&uDAS&3D-NAE&PPNM-AE&MLM-1DAE&MLM-3DAE&~~\\
\rotatebox{90}{ ~~~~ Soil~~~~~~~~~~~ Tree~~~~~~~~~~ Water~~~}
	\includegraphics[trim = 3mm 0mm 0mm 40mm, clip,width=0.11\textwidth]{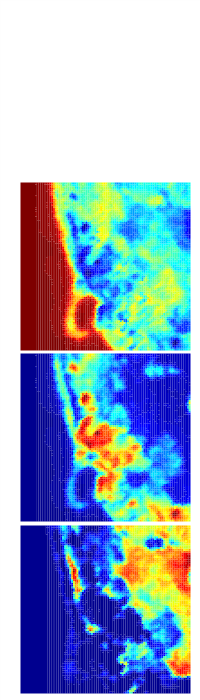} &
	\includegraphics[trim = 3mm 0mm 0mm 40mm, clip,width=0.11\textwidth]{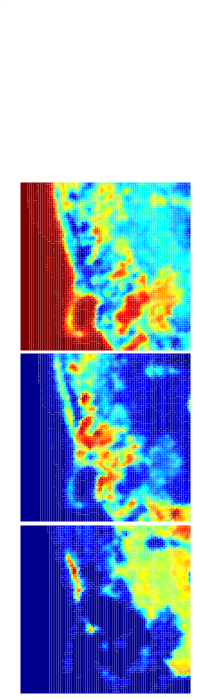}  & 
      \includegraphics[trim = 3mm 0mm 0mm 40mm, clip,width=0.11\textwidth]{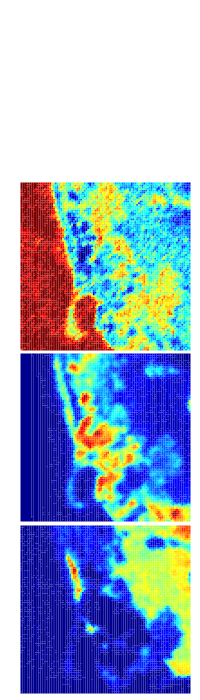} &
	\includegraphics[trim =3mm 0mm 0mm 40mm, clip,width=0.11\textwidth]{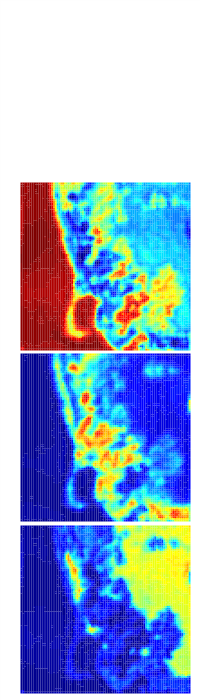} &
	\includegraphics[trim = 3mm 0mm 0mm 40mm, clip,width=0.11\textwidth]{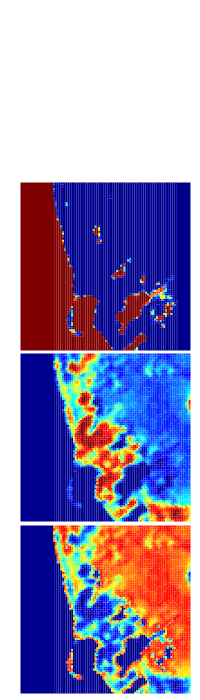}  &
	\includegraphics[trim = 3mm 0mm 0mm 40mm, clip,width=0.11\textwidth]{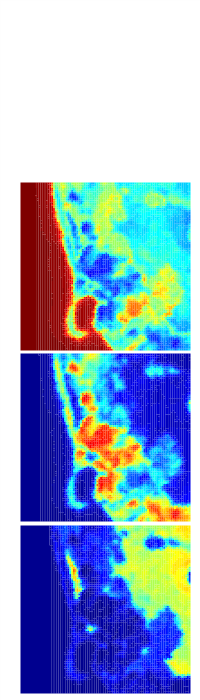}&
	\includegraphics[trim = 3mm 0mm 0mm 40mm, clip,width=0.11\textwidth]{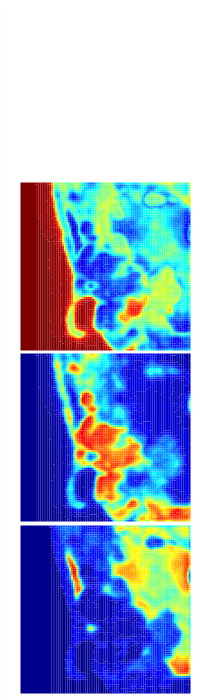} &
\includegraphics[trim =35mm 18mm 0.8mm 8mm, clip,width=0.045\textwidth,height=5.7cm]{colormap.png} 
\end{tabular}
\caption{ Estimated abundance maps for different endmembers on Samson data. Up to bottom: Water, Tree, and Soil. Left to right: VCA+MLM, MLMp, uDAS, 3D-NAE, PPNM-AE, MLM-1DAE, and MLM-3DAE. }  
\label{abun_samson}  
\end{figure*}

\begin{figure*}
	\centering
      \graphicspath{{Figures/}}
	\subfigure[VCA+MLM]{
		\includegraphics[trim = 10mm 10mm 13mm 5mm, clip,width=0.23\textwidth]  {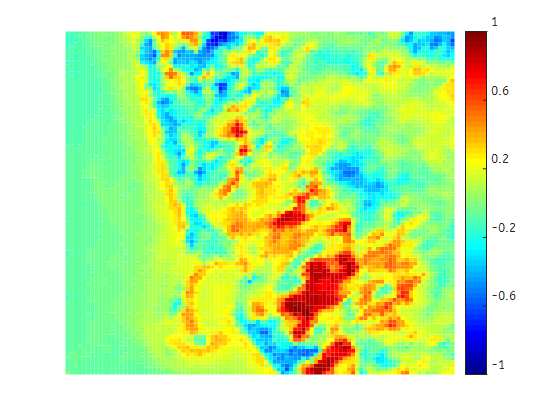}
}
	\subfigure[MLMp]{
		\includegraphics[trim = 10mm 10mm 13mm 5mm, clip,width=0.23\textwidth] {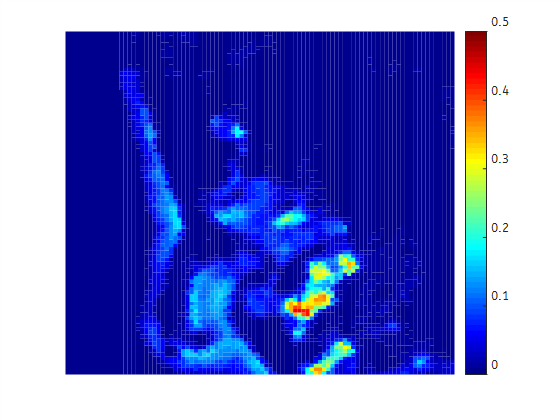}
}
	\subfigure[MLM-1DAE]{
		\includegraphics[trim = 10mm 10mm 13mm 5mm, clip,width=0.23\textwidth]  {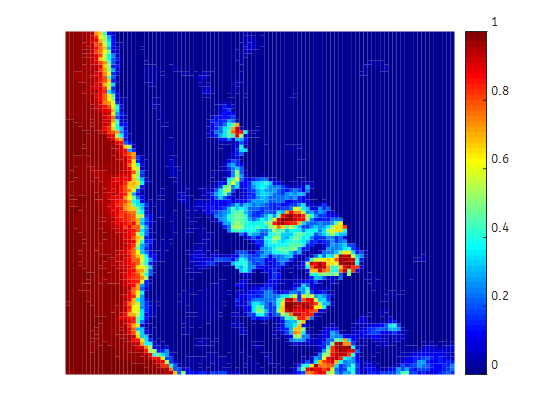}
}
	\subfigure[MLM-3DAE]{
		\includegraphics[trim = 10mm 10mm 13mm 5mm, clip,width=0.23\textwidth]  {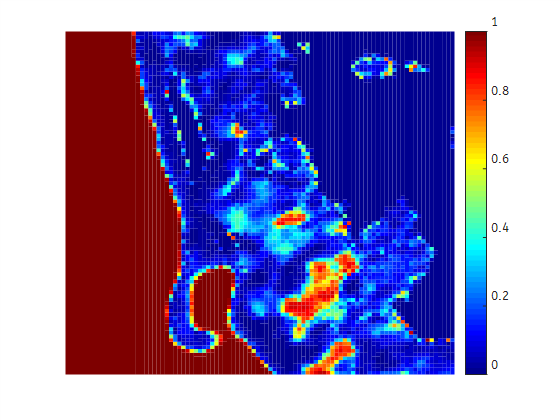}
}
	\caption{\label{samson_p} Visualization of P value by using VCA+MLM, MLMp, MLM-1DAE and MLM-3DAE on Samson data.}
\end{figure*}

\begin{table}\caption{Comparison of pixel averaged SAD ($\times 10^{-2}$) on Samson}
	\centering
	
	\begin{tabular}{c|c}
		\hline
		\hline
		                                                   Method &$$SAD$$\\
		\hline
		$$VCA+MLM$$					      &$\textcircled{\tiny1}~{4.42\pm0.00}$\\
		$$MLMp$$			       &$5.20\pm0.00$\\
		$$uDAS$$			             &$5.14\pm0.00$\\
		$$3D-NAE$$				&$8.34\pm0.57$\\
             $$PPNM-AE$$				&$7.86\pm0.00$\\
		$$MLM-1DAE$$			&$5.00\pm0.10$\\
		$$MLM-3DAE$$			&$\textcircled{\tiny2}4.99\pm0.12$\\
		\hline
		\hline
	\end{tabular}
\label{samsonRE}
\end{table}

\begin{table}\caption{Comparison of pixel averaged SAD ($\times 10^{-2}$) on Urban }
	\centering
	\begin{tabular}{c|c}
		\hline
		\hline
                               Method&SAD\\
		\hline
		$$VCA+MLM$$					&$5.85\pm0.00$\\
		$$MLMp$$				&$5.19\pm0.00$\\
		$$uDAS$$			       &$11.93\pm0.00$\\
		$$3D-NAE$$					&$5.07\pm0.04$\\
       $$PPNM-AE$$			   &$ 13.53\pm0.00$\\
		$$MLM-1DAE$$ 			&$\textcircled{\tiny2}~ 4.29\pm0.02$\\
		$$MLM-3DAE$$			    &$\textcircled{\tiny1}~{4.27\pm0.02}$\\
		\hline
		\hline
	\end{tabular}
\label{urbanRE}
\end{table}

\begin{figure*}  
\centering
\small  
\graphicspath{{Figures/}}
\tabcolsep=0.1mm
\begin{tabular}{cccccccc}
VCA+MLM&MLMp&uDAS&3D-NAE&PPNM-AE&MLM-1DAE&MLM-3DAE&\\
\rotatebox{90}{ ~~~~~ Tree~~~~~~~~~~ Roof~~~~~~~~~ Grass~~~~~~~~~Asphalt}
\includegraphics[trim = 3mm 0mm 0mm 0mm, clip,width=0.11\textwidth]{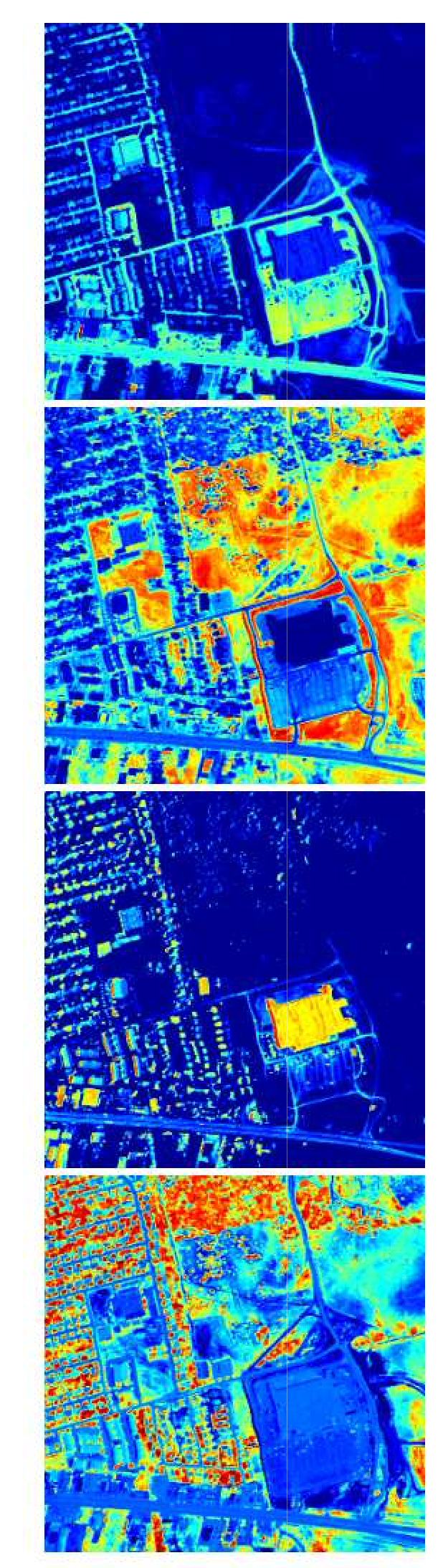} &
	\includegraphics[trim = 3mm 0mm 0mm 0mm, clip,width=0.11\textwidth]{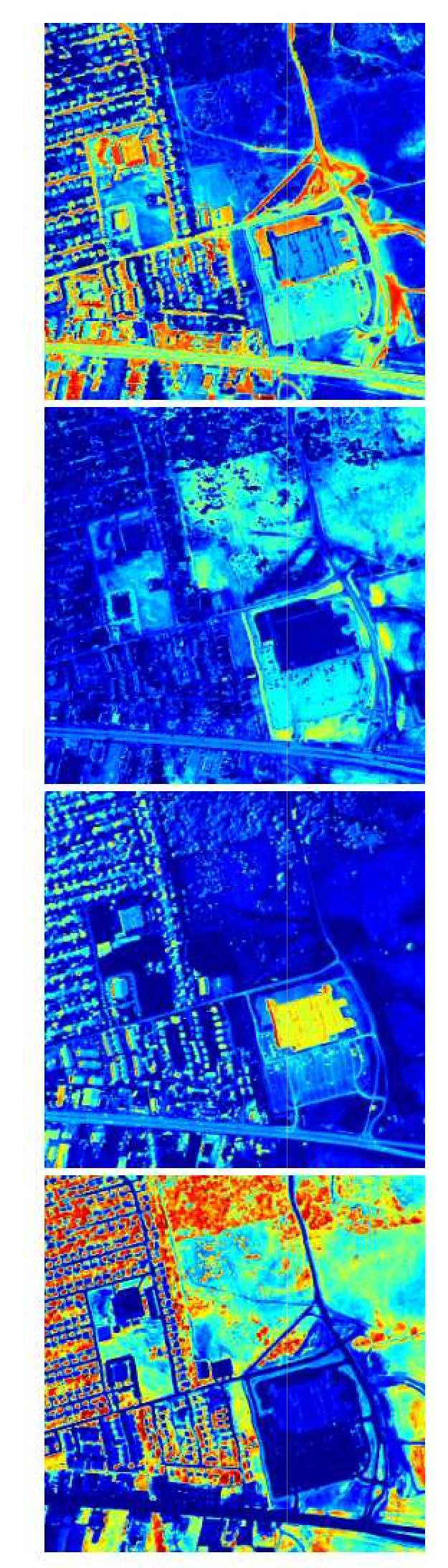}  & 
      \includegraphics[trim = 3mm 0mm 0mm 0mm, clip,width=0.11\textwidth]{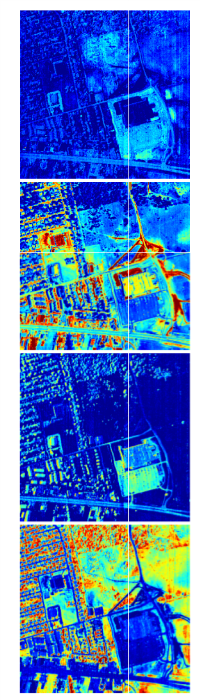} &
	\includegraphics[trim =3mm 0mm 0mm 0mm, clip,width=0.11\textwidth]{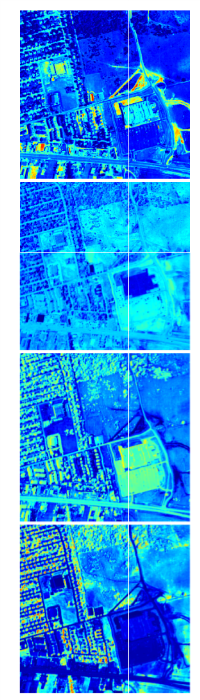} &
	\includegraphics[trim = 3mm 0mm 0mm 0mm, clip,width=0.11\textwidth]{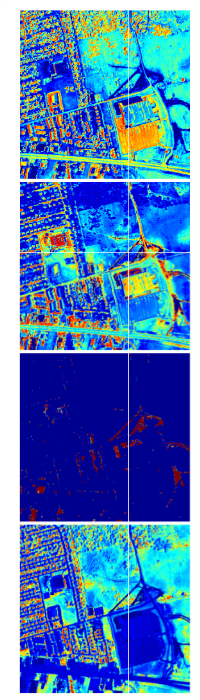}  &
	\includegraphics[trim = 3mm 0mm 0mm 0mm, clip,width=0.11\textwidth]{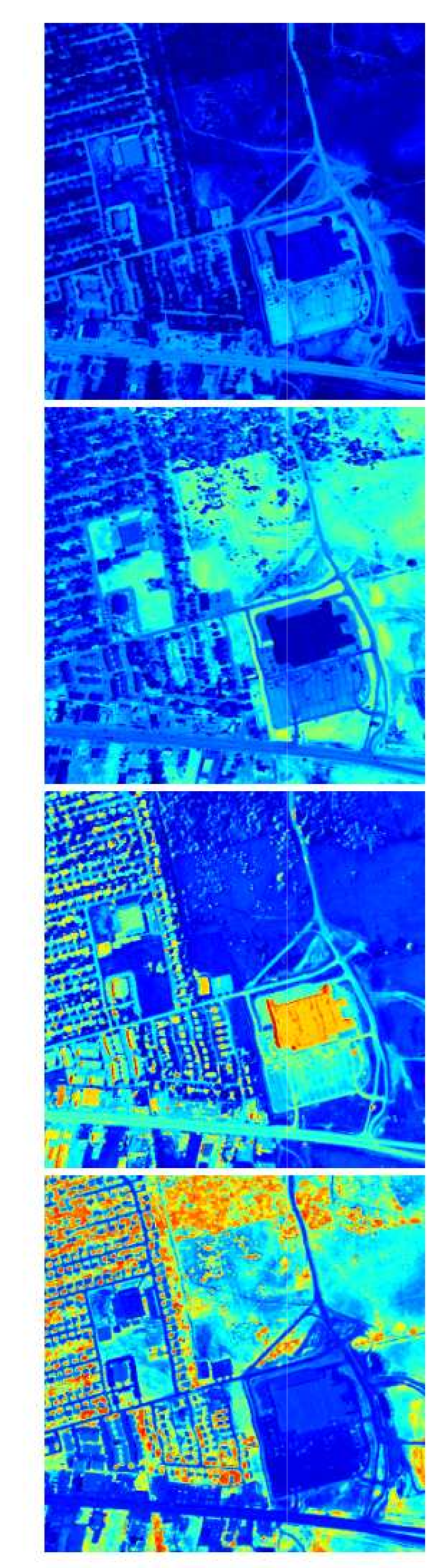}&
	\includegraphics[trim = 3mm 0mm 0mm 0mm, clip,width=0.11\textwidth]{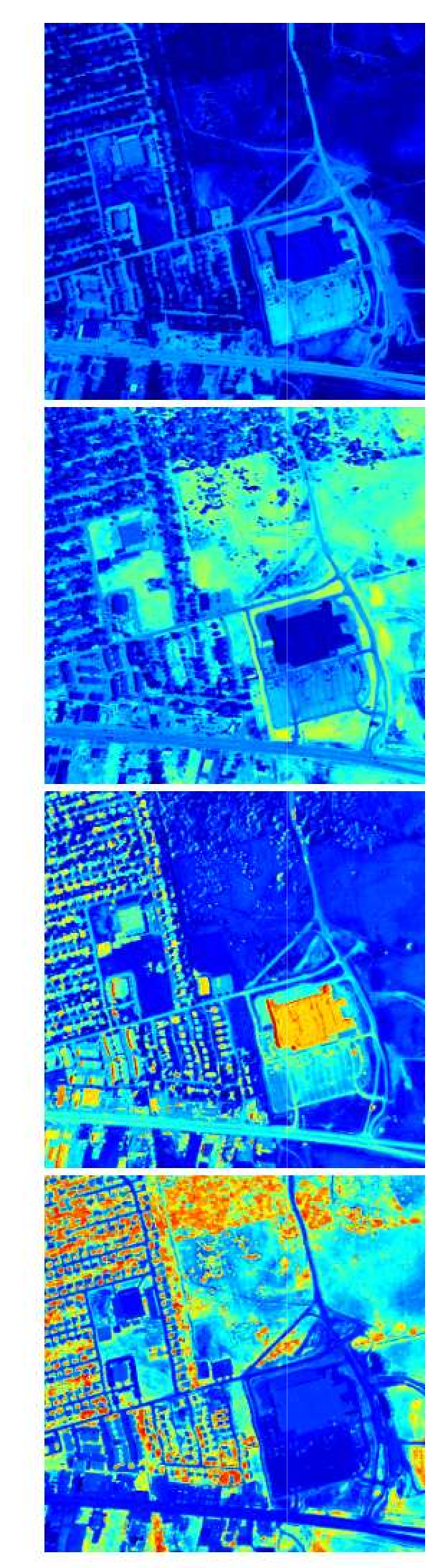} &
\includegraphics[trim =35mm 19mm 0.8mm 8.5mm, clip,width=0.045\textwidth,height=7.55cm]{colormap.png} 
\end{tabular}
\caption{ Estimated abundance maps for different endmembers on Urban data. Up to bottom: Asphalt, Grass, Roof, and Tree. Left to right: VCA+MLM, MLMp, uDAS, 3D-NAE, PPNM-AE, MLM-1DAE, and MLM-3DAE. }  
\label{urban_abun}  
\end{figure*}

\begin{figure*}
	\centering
\graphicspath{{Figures/}}
	\subfigure[VCA+MLM]{
		\includegraphics[trim = 10mm 10mm 14mm 5mm, clip,width=0.23\textwidth]  {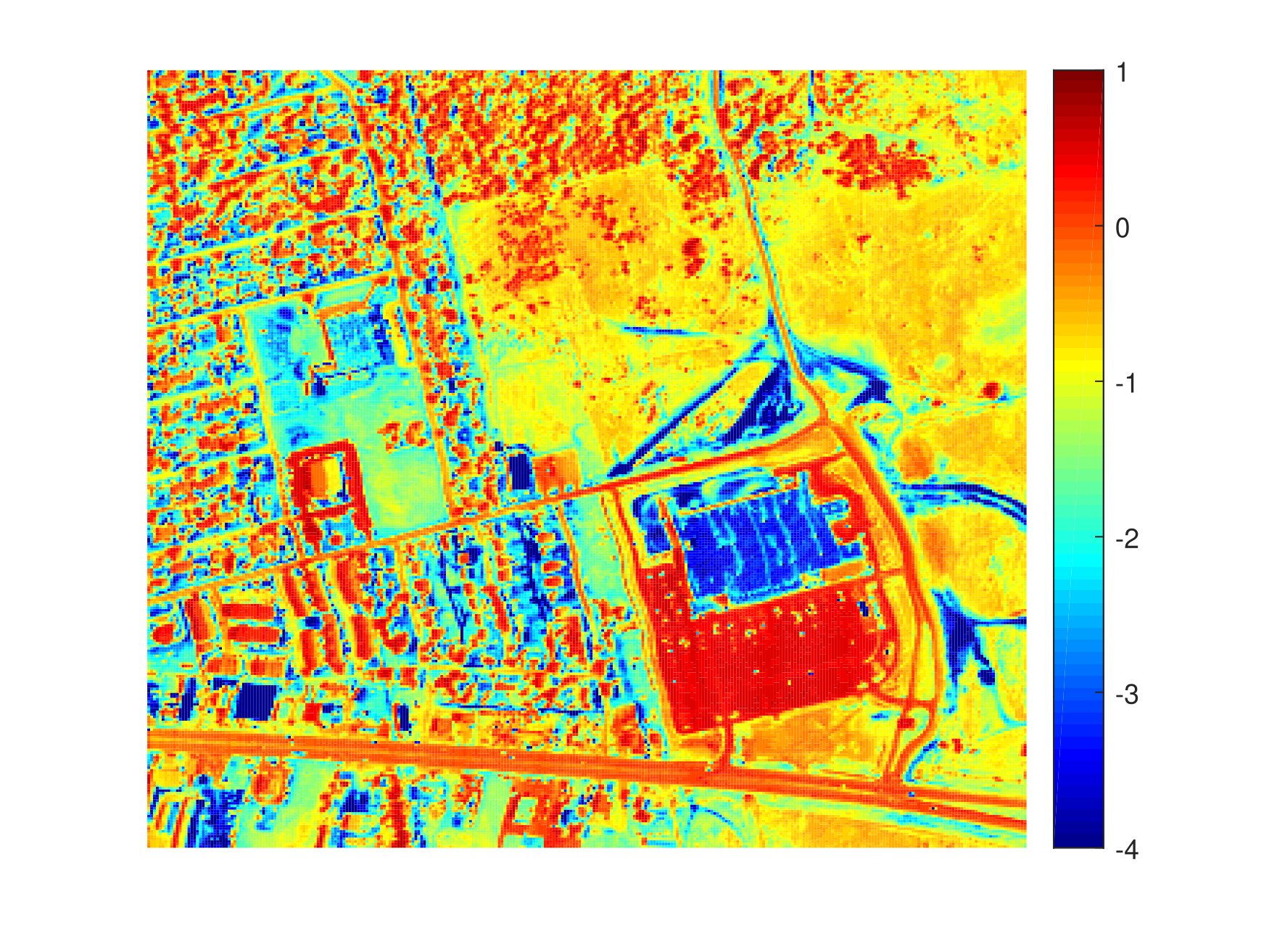}
}
	\subfigure[MLMp]{
		\includegraphics[trim = 10mm 10mm 14mm 5mm, clip,width=0.23\textwidth] {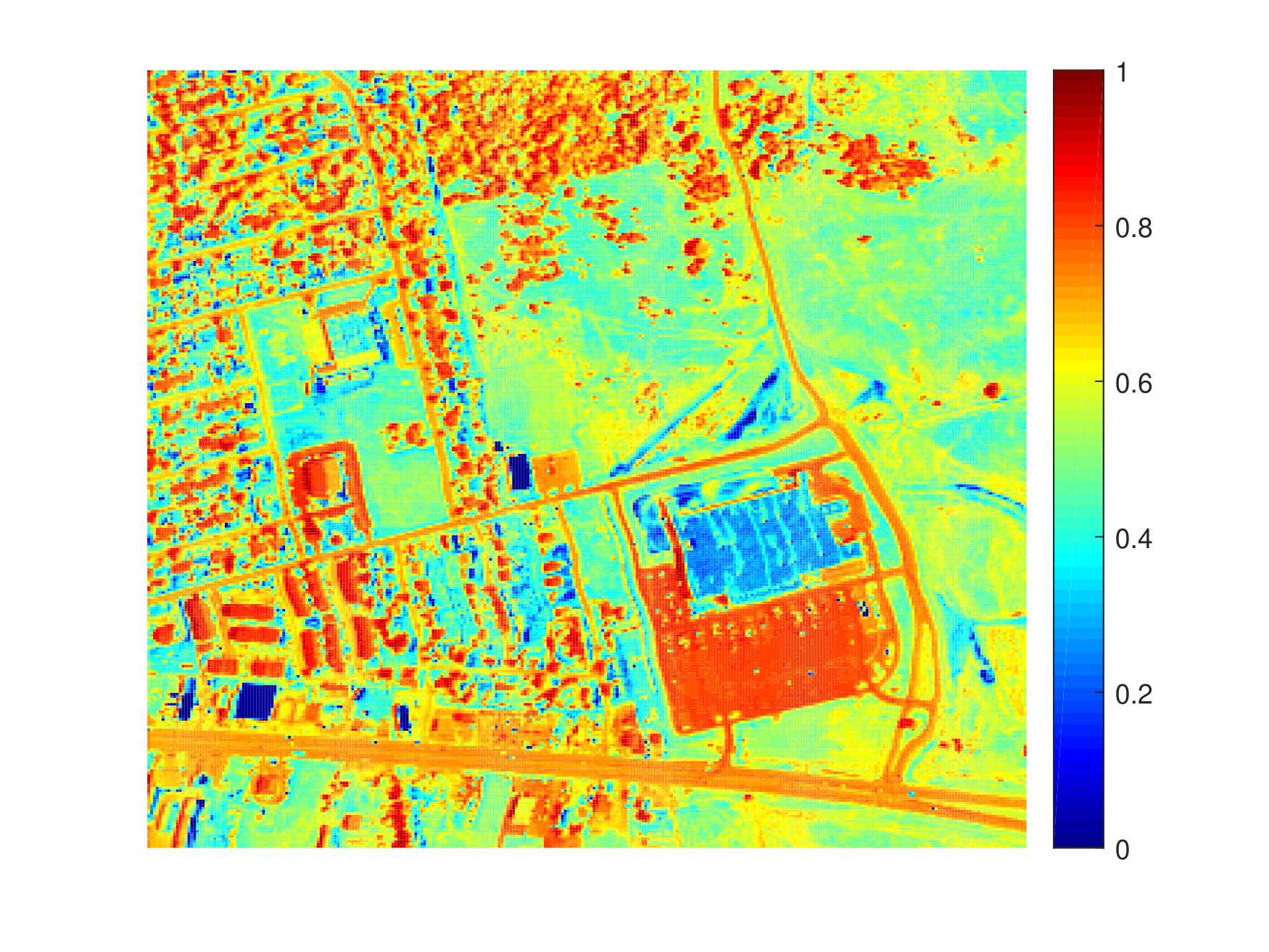}
}
	\subfigure[MLM-1DAE]{
		\includegraphics[trim = 10mm 10mm 14mm 5mm, clip,width=0.23\textwidth]  {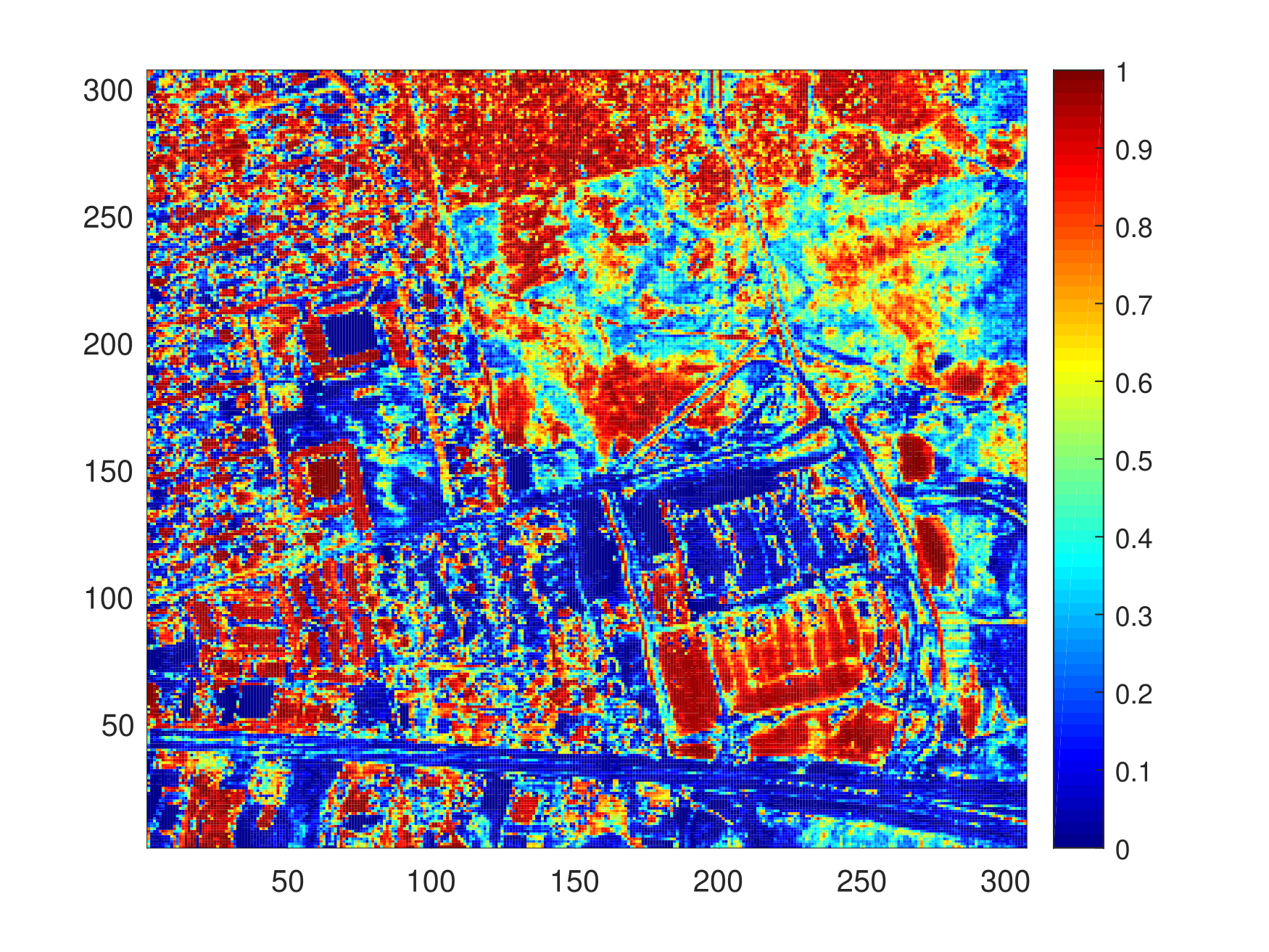}
}
	\subfigure[MLM-3DAE]{
		\includegraphics[trim = 10mm 10mm 14mm 5mm, clip,width=0.23\textwidth]  {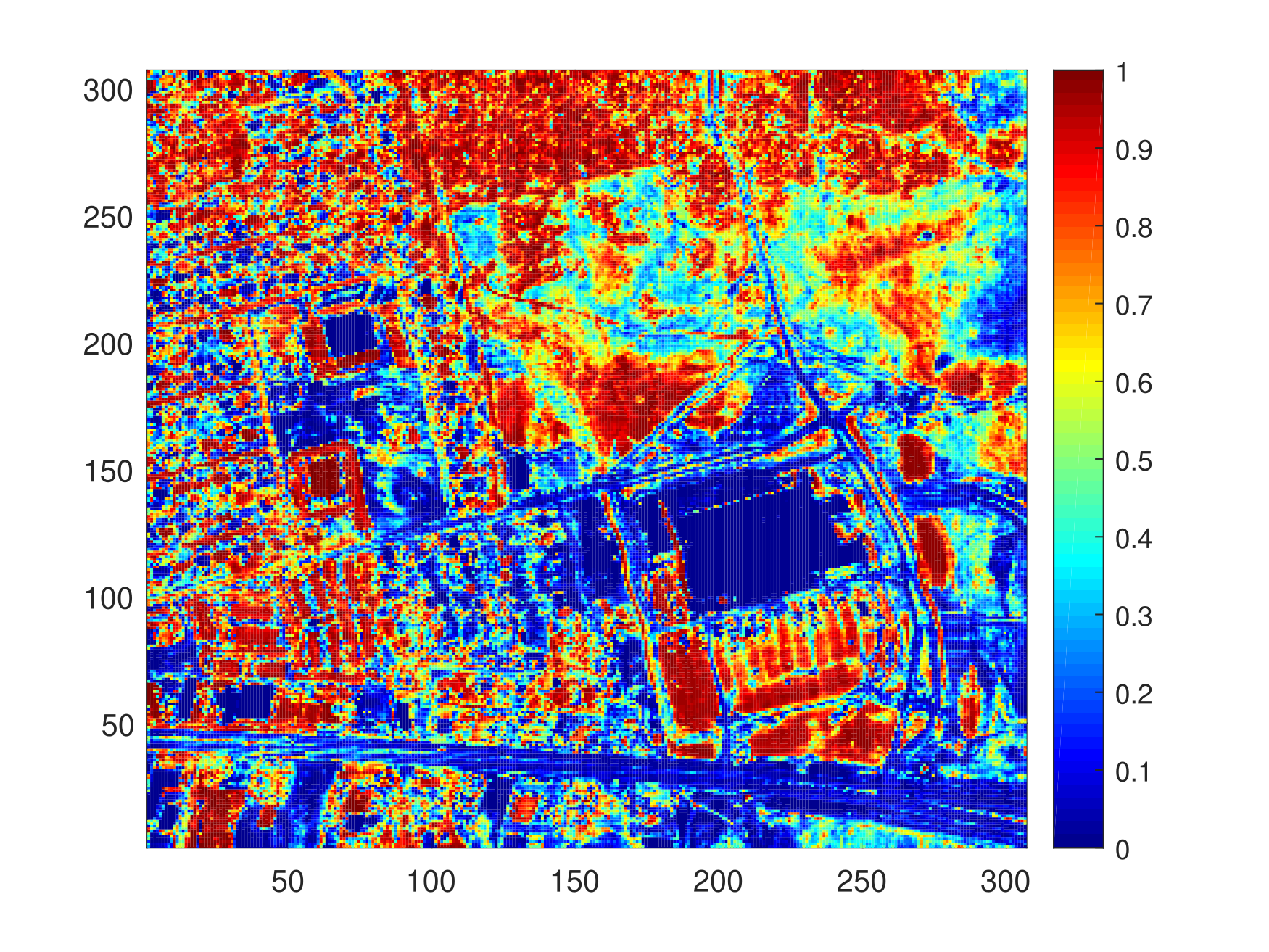}
}
	\caption{\label{urban_p} Visualization of P value by using VCA+MLM, MLMp, MLM-1DAE and MLM-3DAE on Urban data. }
\end{figure*}

\subsection{Real images }
\subsubsection{Samson}

The first real hyperspectral image is the well-known Samson data, with a size of $95\times95$ pixels. After removing the noisy bands, 156 spectral bands are retained covering the wavelength range from 401 to 889 nm. 
There are $R=3$ endmembers in this data, namely Soil, Water, and Tree.

In view of the smaller image scale, the batch size is set to 256 in this data.
The learning rate of the linear decoder in Adam optimizer is set to $0.0005$ with an exponential decay of $0.95$ after every epoch. Meanwhile, the other parts of the network maintained a constant learning rate of $0.0001$. The epoch number is set to 200.

\figurename~\ref{abun_samson} depicts the abundance maps obtained by various algorithms on this data, and \figurename~\ref{samson_p} visualizes the histograms of the estimated P values by different MLM-based methods. 
As the underlying mixing mechanism and groundtruth information are unavailable for real datasets, we provide the pixel SAD values of various algorithms only for reference in~\tablename~\ref{samsonRE}.

\subsubsection{Urban}
The second real data is the Urban data, which is taken over an urban area at Copperas Cove, TX, U, captured by the Hyperspectral Digital Imagery Collection Experiment (HYDICE).
It contains $307 \times 307$ pixels, with 210 bands covering the wavelength range from 400 to 2500 nm. 
Only 162 bands were remained for analysis after removing the channels influenced by water vapor and atmosphere. 
The groundtruth endmembers with $R=4$ is taken as reference, namely the scene is mainly composed of Roof, Asphalt, Tree and Grass.

The batch size is set to 512, and the number of epochs is set to 200. The learning rate of linear decoder in the Adam optimizer is set to $0.0005$ with an exponential decay of $0.95$ after every epoch. Meanwhile, we maintained a constant learning rate of $0.001$ for the rest of the network.
\figurename~\ref{urban_abun} and \figurename~\ref{urban_p} visualize the estimated abundance maps and P-map, respectively. 
\tablename~\ref{urbanRE} compares the pixel SAD by different methods, where the proposed MLM-3DAE is the most competitive. 

\section{conclusion}\label{sec5}
This paper proposed an autoencoder-based unmixing method for MLM, where endmembers, abundance vector, and transition probability parameter were estimated simultaneously.
Taking advantage of elaborate network design, the unmixing process of MLM was realized explicitly.  
The proposed unmixing network had an encoder-decoder architecture: the encoder compressed the input pixel (image patch) into the low-dimensional abundance representation; the decoder successively imitated the linear part and built upon it, the MLM mechanism.    
The transition probability parameter was obtained by using a softmax operation on a high-level feature. 
We considered two modes in the proposed method: MLM-1DAE accounting for only spectral information, and MLM-3DAE incorporating the spectral-spatial information among the neighboring pixels. 
The effectiveness of the proposed method was proved on both the synthetic and real datasets, through a comparative study with the MLM-based solutions and other unmixing networks.

\bibliographystyle{IEEEtran}
\bibliography{bib_fang}
\end{document}